\documentclass[lettersize,journal]{IEEEtran}

\newtheorem{theorem}{Theorem}

\newtheorem{assumption}{Assumption}
\newtheorem{remark}{Remark}
\newtheorem{proof}{Proof}
\usepackage[linesnumbered,ruled,boxed]{algorithm2e}
\usepackage{amsmath,amsfonts}
\usepackage{algorithmic}
\usepackage{array}
\usepackage[caption=false,font=normalsize,labelfont=sf,textfont=sf]{subfig}
\usepackage{textcomp}
\usepackage{stfloats}
\usepackage{adjustbox}
 \usepackage{multirow}
 \usepackage{booktabs}       % professional-quality tables
\usepackage{url}
\usepackage{verbatim}
\usepackage{graphicx}

\def\BibTeX{{\rm B\kern-.05em{\sc i\kern-.025em b}\kern-.08em
    T\kern-.1667em\lower.7ex\hbox{E}\kern-.125emX}}
\usepackage{balance}
\begin{document}
\title{ Topology Learning for Heterogeneous Decentralized Federated Learning over Unreliable D2D Networks}
\author{Zheshun Wu, Zenglin Xu, \IEEEmembership{Senior Member, IEEE}, Dun Zeng, Junfan Li, and Jie Liu, \IEEEmembership{Fellow, IEEE}
\thanks{Copyright (c) 2015 IEEE. Personal use of this material is permitted. However, permission to use this material for any other purposes must be obtained from the IEEE by sending a request to pubs-permissions@ieee.org.}
\thanks{This work was partially supported by a key program of fundamental research from Shenzhen Science and Technology Innovation Commission (No. JCYJ20200109113403826), the Major Key Project of PCL (No. 2022ZD0115301), and an Open Research Project of Zhejiang Lab (NO.2022RC0AB04). (\emph{Corresponding author: Zenglin Xu.})}
\thanks{Zheshun Wu, Zenglin Xu, Junfan Li and Jie Liu are with the School of Computer Science and Technology, Harbin Institute of Technology Shenzhen, Shenzhen 518055, China (e-mail:
wuzhsh23@gmail.com; xuzenglin@hit.edu.cn; lijunfan@hit.edu.cn; jieliu@hit.edu.cn).}
\thanks{Dun Zeng is with the Department of Computer Science and Engineering, University of Electronic Science and Technology of China, Chengdu
611731, China (e-mail: zengdun@std.uestc.edu.cn).}}

\markboth{Journal of \LaTeX\ Class Files,~Vol.~18, No.~9, September~2020}%
{How to Use the IEEEtran \LaTeX \ Templates}

\maketitle

\begin{abstract}
With the proliferation of intelligent mobile devices in wireless device-to-device (D2D) networks, decentralized federated learning (DFL) has attracted significant interest. Compared to centralized federated learning (CFL),  DFL mitigates the risk of central server failures due to communication bottlenecks. However, DFL faces several challenges, such as the severe heterogeneity of data distributions in diverse environments, and the transmission outages and package errors caused by the adoption of the User Datagram Protocol (UDP) in D2D networks. These challenges often degrade the convergence of training DFL models. To address these challenges, we conduct a thorough theoretical convergence analysis for DFL and derive a convergence bound. By defining a novel quantity named unreliable links-aware neighborhood discrepancy in this convergence bound,  we formulate a tractable optimization objective, and develop a novel Topology Learning method considering the Representation Discrepancy and Unreliable Links in DFL, named ToLRDUL. Intensive experiments under both feature skew and label skew settings have validated the effectiveness of our proposed method, demonstrating improved convergence speed and test accuracy, consistent with our theoretical findings.
\end{abstract}

\begin{IEEEkeywords}
Decentralized federated learning, D2D networks, Topology learning, Data heterogeneity, Unreliable links.
\end{IEEEkeywords}

\section{Introduction}
\IEEEPARstart{W}{ith} the growth of mobile devices and edge computing techniques in wireless networks,  and the rapid development of deep learning technology in mobile communication~\cite{DBLP:conf/icc/JiL23,DBLP:journals/vtm/XiongZNDWW19,xiong2018mobile}, federated learning (FL)  has emerged as a  promising solution of distributed edge learning, as it ensures user privacy~\cite{10319759,9944162}.  This paper specifically focuses on the decentralized federated learning (DFL) deployed on device-to-device (D2D) networks. In this system, devices solely exchange parameters with neighboring nodes based on a specific communication topology to facilitate collaborative training, eliminating the need for a central server. Therefore, DFL mitigates the risk of system vulnerability to failures caused by communication bottlenecks at the server~\cite{DBLP:journals/jstsp/YeLL22,DBLP:journals/tnsm/WuWL23}.

\begin{figure}[ht]
\centering
\captionsetup[subfloat]{font=tiny }	
{\includegraphics[width = 0.3\textwidth]{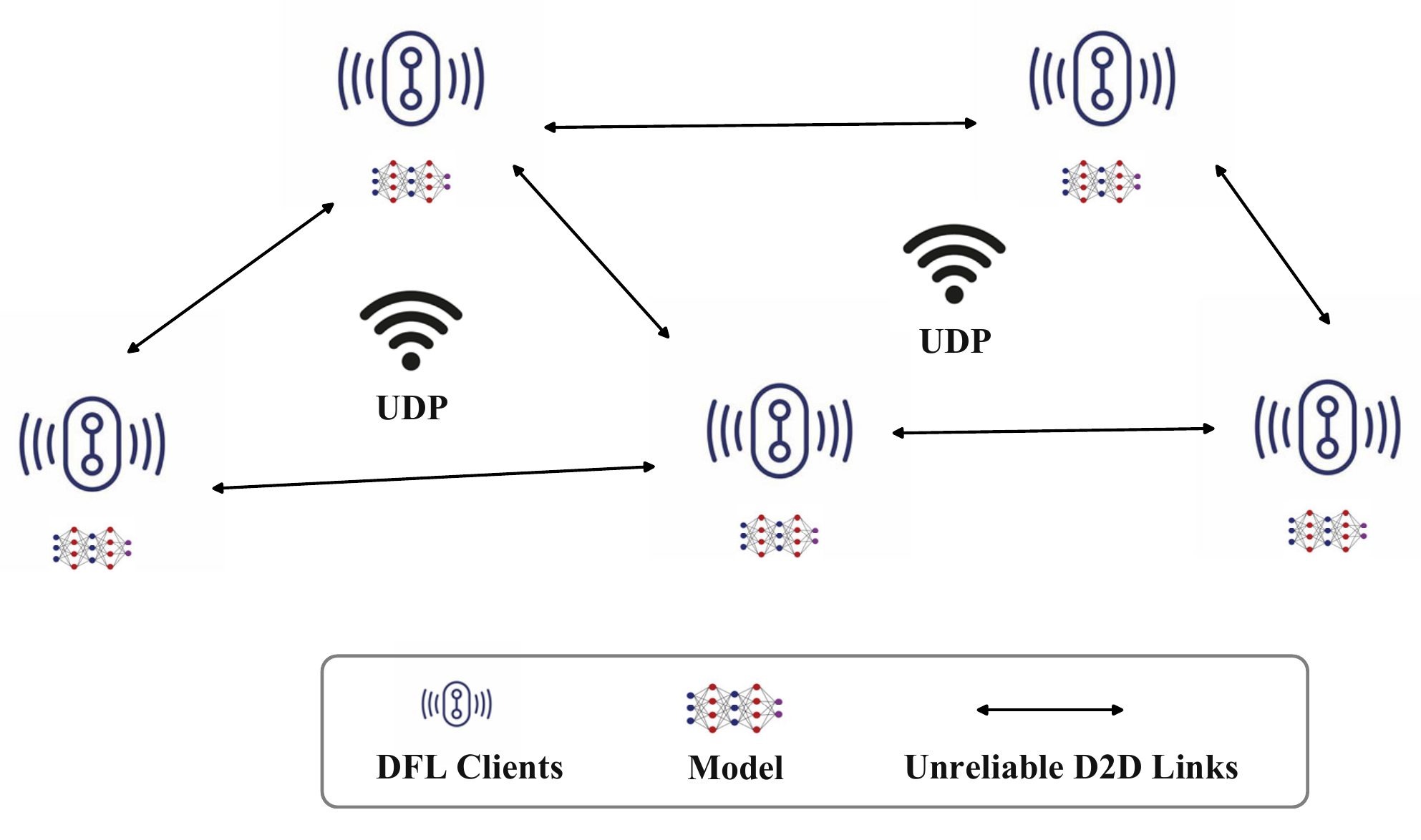}}
\caption{ An illustration of DFL systems deployed in wireless D2D networks. In this system, the sensors serve as the DFL clients. The UDP protocol is used in D2D communication and the D2D links are unreliable.}
\label{fig:topology}
\vspace{-0.58cm}
\end{figure}

However, DFL  encounters several challenges in the training of models. One key issue in DFL is to tackle with data heterogeneity since edge devices often need to collect data from distinct areas to carry out intelligent tasks~\cite{DBLP:conf/sensys/JiX022,DBLP:journals/tccn/XiongZLKNLM21,yan2023convergence}, which leads to serious variations in data distributions~\cite{chen2023exploring,DBLP:conf/aistats/BarsBTLK23,DBLP:journals/tcc/ZengLYZLLN23}.  Another serious challenge in DFL is the transmission outages caused by the User Datagram Protocol (UDP) widely used in D2D networks. Usually, DFL deployed in  D2D networks faces the bottleneck of limited communication resources~\cite{DBLP:journals/tsipn/LiuCZ22}, often relying on the lightweight UDP~\cite{DBLP:journals/jstsp/YeLL22,DBLP:journals/tnsm/WuWL23}. However, the use of UDP compromises reliability, resulting in transmission outages and package errors in inter-device communication~\cite{DBLP:journals/jstsp/YeLL22,DBLP:journals/tnsm/WuWL23,yan2023performance}. Consequently,  parameters shared by DFL clients are affected by both data heterogeneity and unreliable D2D links, further deteriorating the convergence and performance of DFL.

% In other words, the method devised in~\cite{DBLP:journals/jstsp/YeLL22} focused solely on optimizingthe weights of a predefined topology with the fixed set of edges, limiting the ability to overcome the effects of unreliable links. 
Numerous studies have been dedicated to addressing these issues. Some have proposed device scheduling methods~\cite{10319759,9944162,chen2023exploring,DBLP:journals/iotj/LiHYKLXN22,9951138} to alleviate the impact of data heterogeneity. However, these approaches are tailored to CFL with reliable links. In a study outlined in \cite{DBLP:journals/jstsp/YeLL22}, the DFL system deployed in unreliable D2D networks was optimized by adjusting mixing weights to enhance performance. Nevertheless, this approach overlooked the influence of data heterogeneity.  Additionally, the authors of~\cite{DBLP:conf/aistats/BarsBTLK23} optimized the topology based on their proposed quantity termed neighborhood heterogeneity  to enhance the training of DFL models. However, their focus was limited to scenarios with label distribution shift and the deployment of DFL in ideal D2D networks without unreliable links. Additionally, the authors in~\cite{chen2020convergence,yahya2023federated} focus on the convergence time optimization in FL for improving the transmission efficiency, but they also ignore both the unreliable links and data heterogeneity. In summary,  existing research has overlooked the theoretical convergence analysis of DFL  considering both arbitrary data heterogeneity and unreliable D2D links, thus neglecting the enhancement of DFL training based on these crucial factors.

The main contributions of this paper are as follows: (1) We provide a theoretical analysis of   the convergence of DFL with data heterogeneity over unreliable D2D networks. Based on this analysis, we investigate the impact of a new quantity named unreliable links-aware neighborhood discrepancy on the convergence bound of DFL. (2) Motivated by the insights from our theoretical analysis, we develop a novel Topology Learning method considering the Representation Discrepancy and Unreliable Links in DFL, named ToLRDUL, which can enhance the DFL training by minimizing the proposed quantity, further dealing with arbitrary data heterogeneity and unreliable links in DFL.  (3) Extensive  experiments under feature skew and label skew  settings have verified that  ToLRDUL outperforms other baselines in both convergence and test accuracy, which is matched with our theoretical findings.

\begin{comment}

Facing with these challenges, we theoretically analyze the convergence of the data heterogeneity-aware DFL over unreliable D2D networks. Furthermore, we present a novel approach to dynamically learn the  communication topology among  clients by effectively minimizing the derived convergence bound. This approach highlights that accounting for these factors can significantly enhance the performance of D2D-based DFL. 
    We consider the decentralized federated learning (DFL) deployed in bandwidth-limited wireless Device-to-Device (D2D) networks with  mobile users, taking into account the challenge of data heterogeneity. Data collected by  mobile users in D2D networks exhibits spatial heterogeneity, mainly manifested as the  data distribution heterogeneity, bringing challenges to DFL. In addition, due to the restriction of limited bandwidth in the edge, user datagram protocol (UDP), a lightweight transmission protocol with much less communication overhead is more suitable to be adopted in the considered DFL.  However, the delivery is not guaranteed in UDP as the transmitted packages are exposed to unreliable communications and no mechanisms for reliability enhancement like retransmission is implemented. 
\end{comment}

\section{System Model}
In this section, we introduce the details of the considered DFL system. We begin by presenting the decentralized optimization over unreliable D2D networks and then outline the corresponding transmission model. Consistent with~\cite{DBLP:journals/jstsp/YeLL22,DBLP:journals/tnsm/WuWL23,yan2023performance}, we apply the lightweight yet unreliable UDP protocol for DFL, which implies the absence of  reliable transmission mechanisms. Consequently, clients discard received packages if they detect errors in packages using error detection codes.
\subsection{Decentralized optimization over unreliable D2D networks}
In this work, we consider the following decentralized optimization objective over an unreliable D2D network of $N$ clients.
\begin{equation}
    \min_{ \mathbf{w}\in \mathbb{R}^d}[f( \mathbf{w}):=\frac{1}{N}\sum_{i=1}^N f_i(\mathbf{w})],
    \end{equation}
    where $f_i(\mathbf{w}):=\mathbb{E}_{Z\sim P^i_{Z}}[F(\mathbf{w},Z)]$. $\mathbf{w} \in \mathbb{R}^d$ is the model and $F$ is the loss function. $Z=(X,Y)$ is a random variable, where $X$ is the feature and $Y$ is the  label. The data heterogeneity is considered, i.e., $P^i_{Z}\neq P^j_{Z}, \forall i \neq j, i,j \in [N]$ and we hypothesize that clients collect data independently. 

In the considered DFL system, each client maintains its local models $\mathbf{w}^i_t$  and shares the local stochastic gradient $\nabla F_i(\mathbf{w}^i_t):=\nabla F(\mathbf{w}^i_t,Z^i_t)\in \mathbb{R}^{d}$ calculated by data $Z^i_t$ sampled in round $t$ with  neighbouring clients based on a specific communication topology over D2D networks to update models.   Notice that  mobile devices have
the capability to execute edge training with the advancement of edge intelligence  in practical scenarios~\cite{deng2020edge,ren2020accelerating}. Due to the unreliability of the UDP protocol used in D2D networks,  packages may be dropped randomly when received by clients. Following the modeling of~\cite{DBLP:journals/jstsp/YeLL22,DBLP:journals/tnsm/WuWL23},  we assume that the parameters are randomly grouped into multiple packets and transmitted independently. Besides, the aggregation process is affected by the stochastic reception. Specifically, the reception of gradients is modeled using a Bernoulli random vector $\mathbf{m}_{t}^{i,j}\in \mathbb{R}^d$ with its components $\mathbf{m}_{t}^{i,j}(k)\overset{\text{ i.i.d. }}{\sim} Bernoulli(p^{i,j}_t), \forall k \in [d]$.  Here, $\mathbf{m}_{t}^{i,j}(k)=1$ if the $k$-th component of gradients transmitted by client $j$ is successfully received at client $i$, and $\mathbf{m}_{t}^{i,j}(k)=0$ otherwise. $p^{i,j}_t$ thus corresponds to the successful transmission probability  from client $j$ to client $i$ in the $t$-th round. Consequently, the update rule of the $i$-th local model  follows:
\begin{equation}
    \begin{aligned}
        \mathbf{w}_{t+1}^i&=     \mathbf{w}_{t}^i-\eta_{t}\sum_{j=1}^N\theta_{t}^{i,j}\nabla F_j(\mathbf{w}^j_t)\odot \mathbf{m}_{t}^{i,j},\\
    \end{aligned}
\end{equation}
where $\eta_t$ denotes the learning rate and $\odot$ denotes the element-wise multiplication.  The mixing matrix $\boldsymbol{\Theta}_t\in \mathbb{R}^{N\times N}=[\theta^{i,j}_t]_{N \times N}$  of the  topology is a doubly-stochastic matrix, i.e., $\boldsymbol{\Theta}_t^T=\boldsymbol{\Theta}_t, \boldsymbol{\Theta}_t\mathbf{1}=\mathbf{1}$, where $\mathbf{1}\in \mathbb{R}^N$ is an all-ones vector, and $\theta_{t}^{i,j}\neq 0$  only if client $i$ and client $j$ are neighbours.

\subsection{Transmission model of unreliable D2D networks}

In the subsequent analysis, we examine the aforementioned successful transmission probability and the one-round latency for DFL based on the transmission model of the considered D2D network. As shown in Fig.~\ref{fig:topology}, we assume that the sensors collecting environmental data serve as the DFL clients and thus we only consider the terrestrial network and  static channel in this paper. For our analysis, we assume orthogonal
    resource allocation among D2D links~\cite{DBLP:journals/jsac/HuCL23}, and thus define the Signal-to-Noise Ratio (SNR) between two connected D2D clients as $  \mbox{SNR}_{i,j}(t):=(P_{tx}h^{i,j}_t |d^{i,j}_t|^{-2})/\sigma^2$. Here, $P_{tx}$ is the transmit power and $\sigma^2$ is the power of noise. $h^{i,j}_t$ denotes the  Rayleigh fading, modeled as an independent random variable $h^{i,j}_t\sim \exp(1)$ across links and rounds~\cite{DBLP:journals/tnsm/WuWL23}. $|d^{i,j}_t|$ indicates the distance between client $i$ and client $j$. Based on this modeling, we define the successful transmission probability as:
\begin{equation}
    \begin{aligned}
p^{i,j}_t&=\mathbb{P}[   \mbox{SNR}_{i,j}(t)>\gamma_{th}]=\exp(-\frac{\gamma_{th}\sigma^2 }{P_{tx}|d_t^{i,j}|^{-2}})
    \end{aligned}
\end{equation}
where $\gamma_{th}$ denotes the decoding threshold.
\begin{comment}
    \begin{equation}
    \begin{aligned}
p^{i,j}_t&=\mathbb{P}[   \mbox{SNR}_{i,j}(t)>\gamma_{th}]=\mathbb{P}[ h^{i,j}_t > \frac{\gamma_{th}|d^{i,j}_t|^2\sigma^2}{P_{tx}}]\\
&=\exp(-\frac{\gamma_{th}\sigma^2 |d_t^{i,j}|^2}{P_{tx}})
    \end{aligned}
\end{equation}
\end{comment}

In this paper, we focus on the synchronous implementation of DFL. Following the synchronous implementation considered in~\cite{DBLP:journals/tnsm/WuWL23,savazzi2020federated}, we assume that D2D devices exchange signaling with the base station providing D2D links for edge devices to achieve consensus. Hence, the transmission latency in one round   is defined   as:
\begin{equation}
l_{t}:=\max_{i,j\in [N]}\Big[\frac{Q}{B\log(1+ \frac{P_{tx}h^{i,j}_t|d^{i,j}_t|^{-2} }{\sigma^2})}\Big], i\neq j, \theta_{t}^{i,j}\neq 0,
\end{equation}
where $Q$ denotes the size of transmitted packages and $B$ is the bandwidth of each link.

\section{Convergence Analysis}

In this section, we conduct the convergence analysis of   the average model $\Bar{\mathbf{w}}_{t}=\frac{1}{N}\sum_{i=1}^N\mathbf{w}^i_t$ in each round of DFL. Besides, we present a novel neighborhood discrepancy and explore its impact on the derived convergence bound. Our analysis is built upon the following assumptions. 
\begin{assumption}[$\beta$-smoothness]\label{smoothness}
    There exists a constant $\beta >0$ such that for any $Z$, $\forall \mathbf{w}_1,\mathbf{w}_2$, we have $\big\Vert \nabla F(\mathbf{w}_1,Z)- \nabla F(\mathbf{w}_2,Z)\big\Vert\leq \beta \big\Vert \mathbf{w}_1-\mathbf{w}_2\big\Vert$.
\end{assumption}

\begin{assumption}[Bounded variance]\label{variance}
For any $\mathbf{w}$, the variance of the stochastic gradient is bounded by $\xi$, i.e., $\mathbb{E}_{Z \sim P_Z^i}\big\Vert\nabla F(\mathbf{w},Z) -\nabla f_{i}(\mathbf{w})\big\Vert^2 \leq \xi^2, \forall i \in [N]$.
\end{assumption}

\begin{comment}
    \begin{assumption}[Bounded stochastic gradient]\label{Lipschitz}
The expected squared norm of stochastic gradients is uniformly bounded, i.e., $\mathbb{E}\big\Vert \nabla F_i(\mathbf{w})\big\Vert^2\leq L, \forall i \in [N]$.
\end{assumption}
\end{comment}

These assumptions are widely considered in DFL~\cite{DBLP:journals/jstsp/YeLL22,DBLP:conf/aistats/BarsBTLK23,DBLP:journals/tsipn/LiuCZ22}. Next, we introduce a new quantity named unreliable links-aware neighborhood discrepancy and utilize it to derive the convergence bound. Specifically, given $\boldsymbol{\Theta}_t$ and $\mathbf{m}_t^{i,j},\forall i, j \in [N]$, $\bar{H}$ measuring the average expected distance between the oracle average gradient and the aggregated gradients affected by unreliable links is assumed to be bounded as follows.

\begin{assumption}[Bounded unreliable links-aware neighborhood discrepancy] \label{neighborbounded}There exists a constant $\tau >0$, such that  for any $
[\mathbf{w}_t^i]_{i \in [N]}$:
\begin{equation}\label{neighbor} 
   \bar{H}=\frac{1}{N}\sum_{i=1}^N\mathbb{E}\Vert \sum_{j=1}^N\big(\theta_{t}^{i,j}\nabla F_j(\mathbf{w}^j_t)\odot \mathbf{m}_{t}^{i,j}- \frac{1}{N}\nabla F_j(\mathbf{w}^j_t)\big)\Vert^2\leq \tau.
\end{equation}

\begin{remark}
Notice that $ \bar{H}$ can be further bounded by $\frac{1}{N}\sum_{i=1}^N\mathbb{E}\Vert \frac{1}{N}\sum_{j=1}^N\nabla F_j(\mathbf{w}^j_t)- \sum_{j=1}^N\theta_{t}^{i,j}\nabla F_j(\mathbf{w}^j_t)\Vert^2+\frac{1}{N}\sum_{i=1}^N\mathbb{E}\Vert \sum_{j=1}^N\big(\theta_{t}^{i,j}\nabla F_j(\mathbf{w}^j_t)\odot \mathbf{m}_{t}^{i,j}- \theta_{t}^{i,j}\nabla F_j(\mathbf{w}^j_t)\big)\Vert^2$. The first term measures the neighborhood discrepancy by disregarding the influence of unreliable links, which is commonly assumed to be bounded~\cite{DBLP:conf/aistats/BarsBTLK23}. Hence, our paper further focuses on scenarios where the second term capturing the effect of  outages remains bounded to support this assumption. 

Consider a scenario where sensors collecting environmental data function as DFL clients to train models for spatially related intelligent tasks, such as indoor localization. However, these sensors are located far apart, resulting in varying spatial distributions of their data and poorer transmission quality. In short, this assumption implies that the impact of data heterogeneity and unreliable links on the aggregated gradients is governed by $\tau$, which reflects the extent of the deviation between the aggregated gradient and the oracle.
\end{remark}

\end{assumption}
\begin{comment}
  
    \begin{equation}\label{neighbor} 
   \bar{H}=\frac{1}{N}\sum_{i=1}^N\mathbb{E}\big\Vert\frac{1}{N}\sum_{j=1}^N\nabla F_j(\mathbf{w})-\sum_{j=1}^N{\theta}^{i,j}m^{i,j}\nabla F_j(\mathbf{w}) \big\Vert^2\leq \tau.
\end{equation}
  \begin{remark}
 $ \bar{H}$ can be bounded by $\frac{1}{N}\sum_{i=1}^N\mathbb{E}\big\Vert \frac{1}{N} \sum_{j=1}^N\nabla F_j(\mathbf{w})-\sum_{j=1}^N{\theta}^{i,j}\nabla F_j(\mathbf{w}) \big\Vert^2+\frac{1}{N}\sum_{i=1}^N\mathbb{E}\big\Vert \sum_{j=1}^N{\theta}^{i,j}\nabla F_j(\mathbf{w})-\sum_{j=1}^N{\theta}^{i,j}m^{i,j}\nabla F_j(\mathbf{w}) \big\Vert^2$, the first term represents  the discrepancy ignoring the unreliable communications proposed in~\cite{DBLP:conf/aistats/BarsBTLK23} and the second term denotes the effect from transmission outages.
\end{remark}
\end{comment}

Based on these assumptions, we provide the convergence bound of  DFL below and prove it in the appendix.
\begin{theorem} \label{thm1} Let Assumption~\ref{smoothness},~\ref{variance} and~\ref{neighborbounded} hold. We select the stepsize satisfying $\eta_t=\frac{\eta}{\sqrt{T}} \leq \frac{1}{\beta}$, and we have:
 \begin{equation}
     \begin{aligned}
     \min_{0\leq t\leq T}\big\Vert \nabla f(\Bar{\mathbf{w}}_{t}) \big\Vert^2 &\leq     \frac{2(f(\Bar{\mathbf{w}}_{0})- f^*)}{\eta(\sqrt{T}-\beta\eta)}+\frac{\beta\eta+\sqrt{T}}{\sqrt{T}-\beta\eta}\xi^2\\
         &\quad+\frac{(1+\beta^2\eta^2)(\beta\eta+\sqrt{T})}{(\sqrt{T}-\beta\eta)}\tau
     \end{aligned}
 \end{equation}
  where $\eta$ is a constant and  $f^*$ denotes the minimal value of $f$.
\begin{remark}
Theorem~\ref{thm1} demonstrates that as the upper bound $\tau$ of $\bar{H}$ decreases, the convergence of DFL accelerates, which motivates us to develop  a method to learn the  topology $\boldsymbol{\Theta}_t$ of DFL to approximately minimize $\bar{H}$ for improving the training. 
\end{remark}
 \begin{comment}
      \begin{equation}
     \begin{aligned}
         \frac{\mathbb{E}[f_T- f_0]}{T+1}&\leq  \frac{(\beta\eta^2-\eta\sqrt{T})\sum_{t=0}^{T}\big\Vert \nabla f_t \big\Vert^2}{2T(T+1)}+\frac{\beta\eta^2+\eta\sqrt{T}}{2T}\xi^2\\
         &\quad+\frac{\tau(N+\beta^2N+\beta^2)(\beta\eta^2+\eta\sqrt{T})}{2TN}
     \end{aligned}
 \end{equation}
 \begin{remark}
    This upper  bound  is affected by $\tau$, i.e., the discrepancy between the oracle global gradient and the aggregated gradient transmitted over unreliable links  based on the specific communication topology  in practice.  
\end{remark}
 \end{comment}

\end{theorem}

%Based on the analysis in the previous section, this section formulates a tractable optimization objective for topology learning by the upper bound of $\bar{H}$.  We introduce another  assumption widely used in FL~\cite{chen2023exploring,DBLP:journals/jsac/HuCL23} for deriving this bound:

\section{Topology Learning} 

In this section, we derive an upper bound for $\bar{H}$ and formulate a tractable optimization objective for topology learning based on it. Firstly, we introduce another assumption that is widely used in FL~\cite{chen2023exploring,DBLP:journals/jsac/HuCL23} to derive this bound.

\begin{assumption}[Bounded stochastic gradient]\label{Lipschitz}
  The expected norm of stochastic gradients is uniformly bounded by a constant $ L$, i.e., $\mathbb{E}\Vert \nabla F_i(\mathbf{w}^i_{t})\Vert^2\leq L^2, \forall i \in [N].$
\end{assumption}

\begin{theorem}\label{lemm1}Let Assumption~\ref{Lipschitz} hold, $\bar{H}$ can be further bounded as follows.
      \begin{equation}\label{originalobjective}
        \begin{aligned}
   \bar{H}&      \leq\frac{1}{N}\sum_{i=1}^N\mathbb{E}\Vert \sum_{j=1}^N\big(\theta_{t}^{i,j}\nabla F_j(\mathbf{w}^j_t)\odot \mathbf{p}_{t}^{i,j}- \frac{1}{N}\nabla F_j(\mathbf{w}^j_t)\big)\Vert^2\\
&\quad+\frac{dL^2}{N}\sum_{i=1}^N\sum_{j=1}^N(\theta_t^{i,j})^2 p_{i,j}^t(1-p_{i,j}^t).
        \end{aligned}
    \end{equation}
    where $\mathbf{p}_{t}^{i,j}=(p^{i,j}_t,p^{i,j}_t,...,p^{i,j}_t)^T\in\mathbb{R}^{d}$. The proof of this theorem is deferred to the appendix.
\end{theorem}
\begin{remark}
Theorem~\ref{lemm1} reflects that the proposed unreliable links-aware neighbor-
hood discrepancy $\bar{H}$ is impacted by the effect on the aggregated gradients from the expectation of transmission outage, and the variance of this outage.
\end{remark}

\begin{comment}
We begin by analyzing the optimality  of the second and the third term in Eq.~\ref{topobjective} respectively.  Firstly, we set the partial derivative of the second term  to zero, allowing us to identify the  optimal solution as $\theta^{i,j}=\frac{1}{N}, \forall i,j \in [N]$. Moving on to the third term in Eq.~\ref{topobjective}, it drives $\theta^{i,j}, \forall i,j \in [N]$ towards zero, thereby reducing the impact of transmission outages' variance.  Next, we   investigate the optimality of both the second and  third terms. The optimal solution  is given by $\theta^{i,j}=1/[\frac{\lambda_2}{\lambda_1}N(1-p_t^{i,j})+N]$, implying that when $p^{i,j}_t=1$, the corresponding $\theta^{i,j}=1/N$. Besides, a smaller $p^{i,j}_t$ results in a decrease in $\theta^{i,j}$, aligning with our intuition. These observations affirm that these terms are indeed related to the connectivity of topologies.
\end{comment}
      
%We now provide a feasible method for approximating $\{\nabla F(\mathbf{w}_t^i)\}_{i=1}^N$ in Eq.~\eqref{originalobjective}.

 We now present a viable method to approximate the local gradients in Eq.~\eqref{originalobjective} for formulating a tractable optimization objective. Similar to~\cite{DBLP:conf/nips/NguyenTL22}, we employ  a probabilistic network  to generate the representation $\phi \in \mathbb{R}^m (m\ll d)$, with its components  $\phi(k) \sim P_{\phi|Z,W}(k), \forall k\in[m]$, to approximate gradients in Eq.~\eqref{originalobjective}. Utilizing representations (hidden features) produced by models to approximate the associated stochastic gradients is a proper method, as representations encapsulate essential information about training samples~\cite{roberts2022principles}. The representation distributions $\{P^i_{\phi|Z,W}\}_{i=1}^N$ are modeled as Gaussian distributions in this paper. Specifically, for the $i$-th local model, each component  $\phi^i(k)$ of the representation $\phi^i$ used for label prediction is sampled from a Gaussian distribution $\mathcal{N}\big(\phi^i(k);\mu^i_t(k),\sigma^i_t(k)\big)$, where $\mu^i_t(k)$ and $\sigma^i_t(k)$ are the corresponding parameters output by the layer preceding the prediction head.     For approximating the first term on the right-hand side of Eq.~\eqref{originalobjective}, we first define the component-wise average relative entropy for each component $\{\phi^i_t(k)\}_{i=1}^N, \forall k\in [m]$ of representations $\{\phi^i_t\}_{i=1}^N$ as follows.
 \begin{equation}\label{KLterm}
     \hat{H}^k_t(\boldsymbol{\Theta}):=\frac{1}{N}\sum_{i=1}^N D_{kl}\big( \sum_{j=1}^N\theta^{i,j}_t p_{t}^{i,j}\phi^j_t(k)\big\Vert\frac{1}{N}\sum_{j=1}^N\phi^j_t(k) \big)
 \end{equation}
 
The rationale for employing the Gaussian distribution as the representation distribution is that the relative entropy in Eq.~\eqref{KLterm} can be can be computed analytically and we thus avoid using Monte Carlo sampling  methods for compute the term in Eq.~\eqref{KLterm} under other complicated distributions~\cite{DBLP:conf/nips/NguyenTL22}. Therefore, we can analytically calculate $\hat{H}^k_t(\boldsymbol{\Theta})$ using the below expression:
    \begin{equation}\label{bayesian}
\begin{aligned}
\hat{H}_t^k(\boldsymbol{\Theta})&=\frac{1}{N}\sum_{i=1}^N\Big[ \log\frac{\Tilde{\sigma}_t^i(k)}{\Bar{\sigma}_t(k)}+\frac{\frac{1}{N^2}\sum_{j=1}^N (\sigma^j_t(k))^2}{2\sum_{j=1}^N(\theta^{i,j} p_{t}^{i,j})^2(\sigma^j_t(k))^2}\\
&\quad+\frac{(\bar{\mu}_t(k)-\Tilde{\mu}^i_t(k))^2}{2\sum_{j=1}^N(\theta^{i,j} p_{t}^{i,j})^2(\sigma^j_t(k))^2}-\frac{1}{2}\Big]
\end{aligned}
    \end{equation}
where $\bar{\mu}_t(k)=\frac{1}{N}\sum_{j=1}^N\mu^j_t(k)$, $\Tilde{\mu}^i_t(k)=\sum_{j=1}^N{\theta}^{i,j} p_{t}^{i,j}\mu^j_t(k)$, $\bar{\sigma}_t(k)=\frac{1}{N}\sum_{j=1}^N \sigma^j_t(k)$, $\Tilde{\sigma}_t^i(k)=\sum_{j=1}^N\theta^{i,j} p_{t}^{i,j}\sigma^j_t(k)$.  Referring to~\cite{DBLP:conf/nips/NguyenTL22}, we perform the calculation in Eq.~\eqref{bayesian} for each dimension of representations, and sum them across  $m$ dimensions. The sum of $\{\hat{H}_t^k(\boldsymbol{\Theta})\}_{k=1}^m$  serves as an approximation for the term $\frac{1}{N}\sum_{i=1}^N\mathbb{E}\Vert \sum_{j=1}^N\big(\theta_{t}^{i,j}\nabla F_j(\mathbf{w}^j_t)\odot \mathbf{p}_{t}^{i,j}- \frac{1}{N}\nabla F_j(\mathbf{w}^j_t)\big)\Vert^2$ in Eq.~\eqref{originalobjective}, which is justified since relative entropy is well-suited for measuring distribution discrepancy.

Given that the value of    $L$  is unknown in practical scenarios,  we replace the term $L^2$ in Eq.~\eqref{originalobjective} with a hyperparameter $\lambda>0$. Based on the aforementioned results, we formulate the optimization objective for topology learning as follows:
        \begin{equation}\label{topobjective}
\begin{aligned}
&\min_{\boldsymbol{\Theta} \in \mathcal{S}} \Big\{ g_t(\boldsymbol{\Theta}):= \sum_{k=1}^m\hat{H}^k_t(\boldsymbol{\Theta})+\frac{d\lambda}{N}\sum_{i,j=1}^N(\theta^{i,j})^2p_{t}^{i,j}(1-p_{t}^{i,j})\Big\},
\end{aligned}
    \end{equation}
    where $\mathcal{S}$ is a set of doubly-stochastic matrix. The underlying idea of this optimization objective is to learn a topology both enabling aggregated representations affected by outages to approximate the oracle average representation, and mitigating the influence from the variance of unreliable links.

\begin{algorithm}[h]
 \KwIn{ Initialization $\boldsymbol{\Theta}_0=\mathbf{I}$ and hyperparameters  $\lambda\geq 0,0\leq \gamma\leq 1, 1\leq K<T, K \in \mathbb{N}^{+}$.}
  \For{round  $t \in [T]$ }{
  
  \For{client $i \in [N]$ in parallel}{
  $\boldsymbol{\Theta}_t=\boldsymbol{\Theta}_{t-1}$\;
    \If{$t$ mod $K$=0}{  
 Exchange encrypted information to attain  $(\boldsymbol{\mu}_t,\boldsymbol{\sigma}_t)$ and $ \mathbf{P}_{t}$ via all-reduce\;
  $\Tilde{\boldsymbol{\Theta}}=\arg\min_{\mathbf{S}\in \mathcal{S}}<\mathbf{S},\nabla g_t(\boldsymbol{\Theta}_{t-1})>$\;
$\boldsymbol{\Theta}_t=(1-\gamma)\boldsymbol{\Theta}_{t-1}+\gamma\Tilde{\boldsymbol{\Theta}} $\;
} 
  
   $ \mathbf{w}_{t+1}^i=     \mathbf{w}_{t}^i-\eta_{t}\sum_{j=1}^N\theta_{t}^{i,j}\nabla F( \mathbf{w}_{t}^j,Z_t^j)\odot \mathbf{m}_{t}^{i,j}$\;   
  }
} 
  \caption{Topology Learning considering the Representation Discrepancy and Unreliable Links for DFL}
  \label{alg1}
\end{algorithm}

Drawing from the method outlined in~\cite{DBLP:conf/aistats/BarsBTLK23}, we adopt the Frank-Wolfe (FW) algorithm to find the approximate solution of Eq.~\eqref{topobjective}, which is well-suited for learning a sparse parameter over convex hulls of finite set of atoms~\cite{DBLP:conf/aistats/BarsBTLK23}. In our context,  $\mathcal{S}$ corresponds to the convex hull of the set  of all permutation matrices~\cite{DBLP:conf/aistats/BarsBTLK23}, enabling  our approach to  learn a sparse topology including both edges and their associated
mixing weights. To execute this approach, clients encrypt and exchange packages every $K$ round. These packages encompass locations used for calculating $\mathbf{P}_t=[p_{t}^{i,j}]_{N\times N}$, as well as $(\boldsymbol{\mu}_t,\boldsymbol{\sigma}_t)$ denoting the parameters of all the Gaussian distributions. This scheme incurs low communication costs since the lower dimension of representations compared to gradients and lower communication frequency. Clients can thus utilize reliable protocols to exchange these packages without experiencing outages. The detailed workflow of this method is described in Algorithm~\ref{alg1}.

\section{Numerical Results}   
In this section, we conduct numerical experiments to evaluate the effectiveness of the proposed ToLRDUL. We provide the details of the transmission model used in the considered D2D networks below. The  bandwidth  is set to $5$MHz, the transmission power $P_{tx}$  is set to $10$dBm and the noise power $\sigma^2$  is set to $-169$dBm. The decoding threshold $\gamma_{th}$ is set to $0$dB.  We deploy the D2D network in a  region of $1000 \times 1000 $ $m^2$ with $100$ randomly distributed clients participating in DFL. The size of transmitted packages is set to $1.2$MB.

 We utilize CNN as the backbone model. Clients perform SGD  with a mini-batch size of $128$ and a learning rate of $0.1$. The local epoch is set to $1$ and the learning rate decay is set to $0.996$. We employ two widely used variants of the CIFAR-10 dataset: Dirichlet CIFAR-10 and Rotated CIFAR-10~\cite{de2022mitigating}. These two variants are chosen to model label skew and feature skew settings respectively. For Dirichlet CIFAR-10, we set the parameter of Dirichlet distribution to $0.1$. For Rotated CIFAR-10, we rotate images clockwise by $3.6^{\circ}$ for each time and distribute them to $100$ clients one by one. We test the average model $\Bar{\mathbf{w}}_t$ on the global test set following the same distribution as that of all the clients. In addition, we calculate the global training loss by averaging the local training losses from different clients, and regard it as the metric for convergence.  We compare our method with STL-FW proposed in~\cite{DBLP:conf/aistats/BarsBTLK23},  which only leverages the label distribution shift to learn the topology.  Specifically, STL-FW requires  clients to collect the class proportions $\pi_{ik}=P_{i}(Y=k)$ for each client $i$ and each class $k$ and use Frank-Wolfe method to solve the optimization problem $\min_{\boldsymbol{\Theta}\in \mathcal{S}}\{\frac{1}{n}\Vert \boldsymbol{\Theta}\Pi-\frac{\mathbf{1}\mathbf{1}^T}{n}\Pi\Vert_F^2+\frac{\lambda}{n}\Vert\boldsymbol{\Theta}- \frac{\mathbf{1}\mathbf{1}^T}{n}\Vert_F^2\}$, where $\Pi \in[0,1]^{n\times K}$ contains the class proportions $\{\pi_{ik}\}$. Besides, we set the random $r$-regular graph~\cite{DBLP:conf/aistats/BarsBTLK23} and  fully-connected graph  as  baselines, with uniform weights for all activated links. The hyperparameter  $\lambda$ and $K$ of ToLRDUL are set to $0.001$ and $3$ respectively. 
                \begin{figure}[ht]
          
\captionsetup[subfloat]{font=scriptsize}	
\centering
   	\subfloat[ Training loss]{\includegraphics[width = 0.26\textwidth]{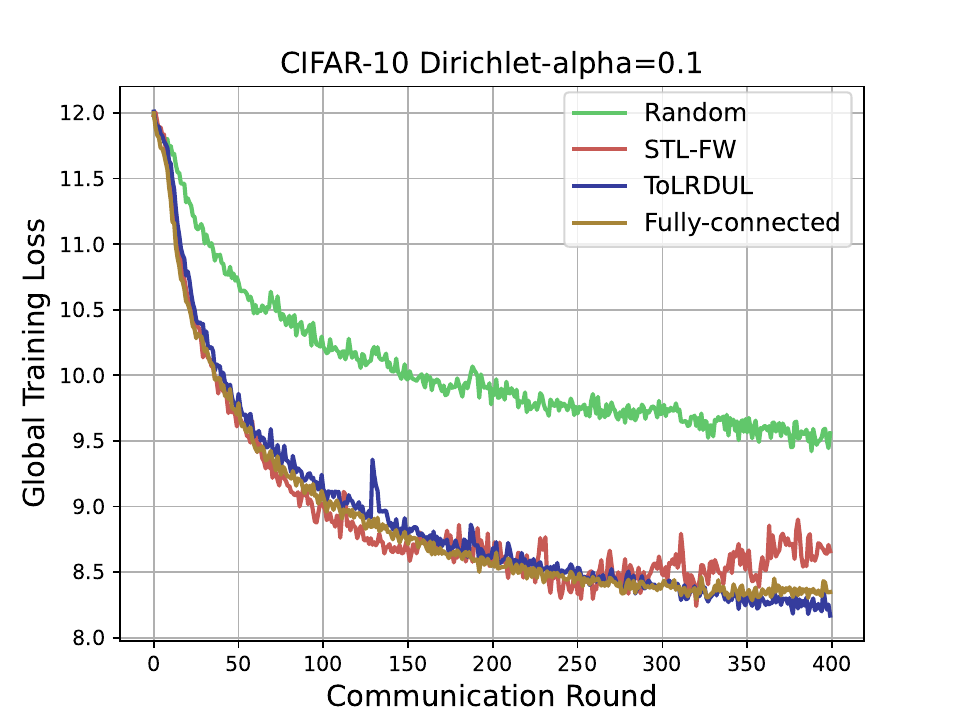}}
    \subfloat[ Test accuracy ]{\includegraphics[width = 0.26\textwidth]{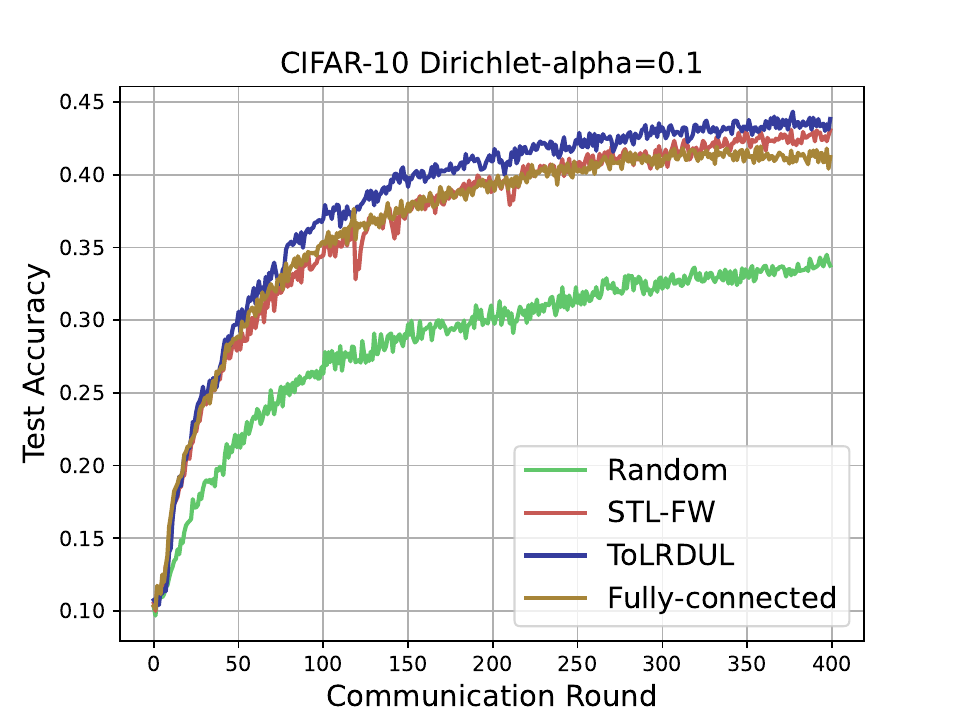}}
    
   	\subfloat[ Training loss]{\includegraphics[width = 0.26\textwidth]{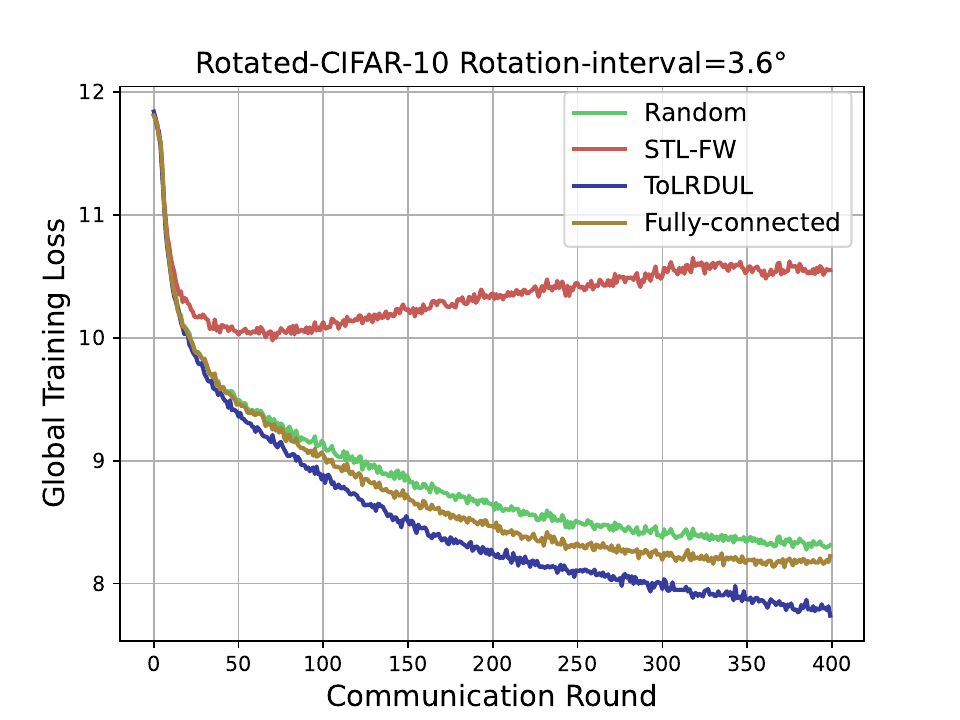}}
        \subfloat[ Test accuracy ]{\includegraphics[width = 0.26\textwidth]{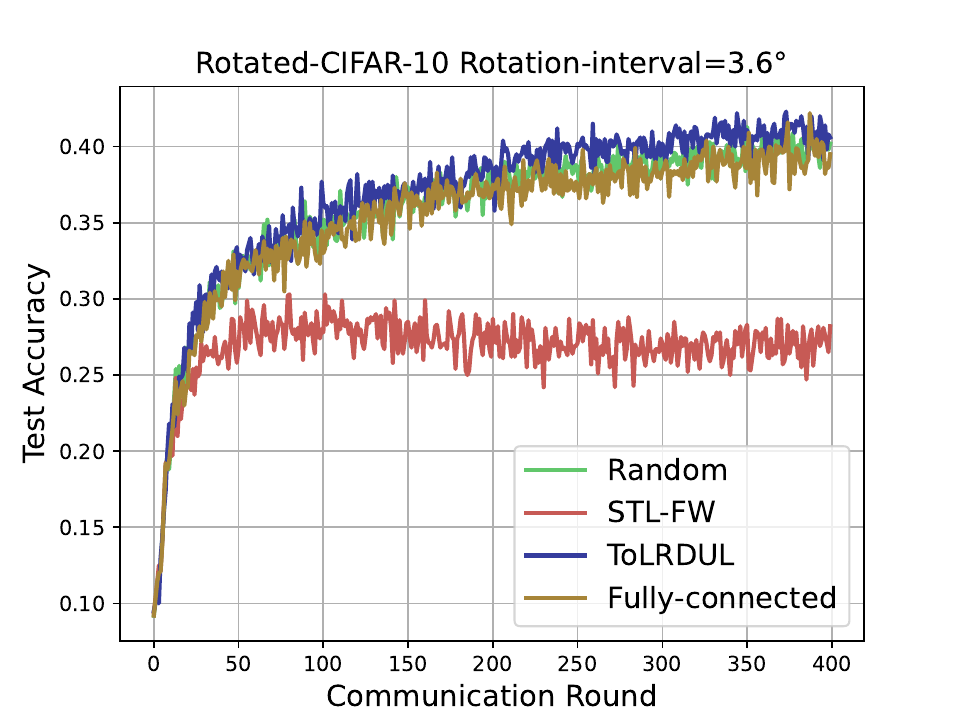}}
\caption{Convergence analysis of DFL with ToLRDUL and other baselines on label-skew and feature skew  CIFAR-10.}
\label{fig:dirichlet}
\end{figure}

We begin by analyzing the convergence of ToLRDUL and other baselines. We fix the degree  $r$ of nodes to be $2$ in this analysis.  As shown in Fig.~\ref{fig:dirichlet} (a) and (b), ToLRDUL demonstrates faster and more stable convergence, achieving the highest test accuracy on Dirichlet CIFAR-10  after $400$ rounds. On the contrary, STL-FW neglects the impact of unreliable links, resulting in unstable convergence. Furthermore, the fully-connected graph performs worse than ToLRDUL due to an excessive number of unreliable links among connected clients. This leads to frequent transmission outages, which hinder the training of the model. Next, we turn our attention to the convergence analysis on Rotated CIFAR-10 in Fig.~\ref{fig:dirichlet} (c) and (d). Similarly, ToLRDUL outperforms baselines in both convergence and test accuracy. Notably, STL-FW performs even worse than the random $r$-regular graph on Rotated CIFAR-10 since it fails to employ the feature distribution discrepancy to learn the topology when confronted with the challenge of feature distribution skew.

\begin{table}[h]

\caption{Dirichlet CIFAR-10 after $300$ communication rounds}
    \label{tab:table1}
    \centering
\begin{adjustbox}{width=0.85\width}
\begin{tabular}{clcccccc}
\toprule
\multicolumn{2}{c}{\multirow{2}{*}{Method}} & \multicolumn{3}{c}{Test accuracy}       & \multicolumn{3}{c}{Transmission latency} \\ %\cline{3-6} 
\cmidrule(r){3-5}
\cmidrule(r){6-8} 
\multicolumn{2}{c}{}                 & $r$=2         & $r$=4    & \multicolumn{1}{c}{$r$=10 }  & $r$=2    & $r$=4            & $r$=10         \\ \midrule

\multicolumn{2}{c}{ToLRDUL (ours)}                 &\textbf{43.42\%} & \textbf{43.43\%}&\multicolumn{1}{c}{\textbf{43.77\%}} &  \textbf{44.88s}  & \textbf{56.46s}  & \textbf{65.73s}      \\ %\hline
\multicolumn{2}{c}{STL-FW}                & 41.10\%  & 42.88\%&\multicolumn{1}{c}{41.92\%} & 62.38s    &  74.45s   &  85.70s       \\ %\hline
\multicolumn{2}{c}{Random}                & 33.16\% & 42.61\%&\multicolumn{1}{c}{41.77\%} &  61.83s        &  73.62s&85.30s     \\ %\hline
\multicolumn{2}{c}{Fully-connected}                & 41.61\% &41.61\%& \multicolumn{1}{c}{41.61\%} & 92.13s     & 92.13s    & 92.13s      \\ \bottomrule
\end{tabular}
\end{adjustbox}
\end{table}

\begin{table}[h]

\caption{Rotated CIFAR-10 after $300$ communication rounds}
    \label{tab:table2}
    \centering
\begin{adjustbox}{width=0.85\width}
\begin{tabular}{clcccccc}
\toprule
\multicolumn{2}{c}{\multirow{2}{*}{Method}} & \multicolumn{3}{c}{Test accuracy}       & \multicolumn{3}{c}{Transmission latency} \\ %\cline{3-6} 
\cmidrule(r){3-5}
\cmidrule(r){6-8} 
\multicolumn{2}{c}{}                 & $r$=2         & $r$=4    & \multicolumn{1}{c}{$r$=10 }  & $r$=2    & $r$=4            & $r$=10         \\ \midrule

\multicolumn{2}{c}{ToLRDUL (ours)}                 &\textbf{41.40\%} & \textbf{41.70\%}&\multicolumn{1}{c}{\textbf{41.50\%}} &  \textbf{45.23s}  & \textbf{56.69s}  & \textbf{66.22s}      \\ %\hline
\multicolumn{2}{c}{STL-FW}                & 30.30\%  & 31.70\%&\multicolumn{1}{c}{40.60\%} &  59.80s    &  71.80s   &  82.34s       \\ %\hline
\multicolumn{2}{c}{Random}                & 40.20\% & 41.20\%&\multicolumn{1}{c}{40.50\%} &  60.81s        &  72.52s&83.34s     \\ %\hline
\multicolumn{2}{c}{Fully-connected}                & 39.10\% &39.10\%& \multicolumn{1}{c}{39.10\%} & 90.50s     & 90.50s    & 90.50s      \\ \bottomrule

\end{tabular}

\end{adjustbox}
\end{table}

Then we investigate the effect of the degree $r$ of nodes on two variants of CIFAR-10. The results are summarized  in Table~\ref{tab:table1} and Table~\ref{tab:table2}, where the unit of transmission delay is seconds.  We find that increasing the value of $r$   leads to an enlargement of the transmission latency $l_t$ since activating more links may result in a higher number of unreliable links. Denser topologies result in larger values of $l_t$,  aligning with our intuition. Besides, an excess of activated links in fully-connected graph cause severe package errors and thus degrade the test accuracy.  Notice that ToLRDUL achieves the lowest latency and the highest test accuracy under all three  settings since it utilizes both data heterogeneity and channel conditions to learn the topology for performing  DFL better.

\section{Conclusion}
In this paper, we present a theoretical convergence analysis for DFL with data heterogeneity over unreliable D2D networks. Furthermore, we introduce a novel quantity called unreliable links-aware neighborhood discrepancy and show its influence on the derived convergence bound. Building upon this observation, we develop a topology learning method named ToLRDUL to  approximately minimize the proposed quantity for improving the convergence and performance of DFL.  To validate the effectiveness of our approach, we conduct extensive experiments that confirm its superiority over other baselines, in accordance with our theoretical findings.

\appendix
\subsection{Proof of Theorem~\ref{thm1}}
\begin{proof}

\begin{comment}
= -\eta_t\big\Vert \nabla f(\bar{\mathbf{w}}_{t}) \big\Vert^2+\eta_t\mathbb{E}\big< \nabla f(\bar{\mathbf{w}}_{t}),\nabla f(\bar{\mathbf{w}}_{t})-\sum_{i,j=1}^N\frac{\Tilde{\theta}^{i,j}_t}{N}\nabla F_j(\mathbf{w}_t^j)\big>\\
     = -\eta_t\big\Vert\nabla f(\bar{\mathbf{w}}_{t}) \big\Vert^2+\mathbb{E}\big< \nabla f(\bar{\mathbf{w}}_{t}), \eta_t\nabla f(\bar{\mathbf{w}}_{t})-\frac{\eta_t}{N}\sum_{i,j=1}^N\Tilde{\theta}^{i,j}_t\nabla F_j(\mathbf{w}_t^j)\big>\\
\end{comment}

According to Assumption~\ref{smoothness}, we have:
        \begin{equation}
    \begin{aligned}
            f(\bar{\mathbf{w}}_{t+1})
& \leq f(\bar{\mathbf{w}}_{t})\underbrace{-\mathbb{E}\big< \nabla f(\bar{\mathbf{w}}_{t}),\sum_{i,j=1}^N\frac{\eta_t\theta^{i,j}_t}{N}\nabla F_j(\mathbf{w}_t^j)\odot\mathbf{m}_t^{i,j}\big>}_{T_1}\\
& \quad+\underbrace{\frac{\beta}{2}\mathbb{E}\big\Vert \frac{\eta_t}{N}\sum_{i,j=1}^N\theta^{i,j}_t\nabla F_j(\mathbf{w}_t^j)\odot\mathbf{m}_t^{i,j}\big\Vert^2}_{T_2}.
        \end{aligned}
    \end{equation}

\begin{comment}
          &\leq -\frac{\eta_t}{2}\big\Vert \nabla f(\bar{\mathbf{w}}_{t}) \big\Vert^2+\frac{\eta_t}{2N}\mathbb{E}\big\Vert \nabla  f(\mathbf{\bar{W}}_t)-\boldsymbol{\Tilde{\Theta}_t}\nabla F(\mathbf{W}_t) \big\Vert^2_F\\
\end{comment}
Next, we proceed to add and subtract the term $\eta_t\nabla f(\bar{\mathbf{w}}_{t})$ in $T_1$, and we can thus convert $T_1$ into: $-\eta_t\big\Vert \nabla f(\bar{\mathbf{w}}_{t}) \big\Vert^2+\frac{\eta_t}{N}\mathbb{E}\sum_{i=1}^N\big<\nabla f(\bar{\mathbf{w}}_{t}),\nabla f(\bar{\mathbf{w}}_{t})-\sum_{j=1}^N\theta^{i,j}_t\nabla F_j(\mathbf{w}_t^j)\odot\mathbf{m}_t^{i,j}\big>$.

We then upper-bound this term  for  bounding $T_1$:
 \begin{equation}
     \begin{aligned} 
   T_1
    &\leq -\frac{\eta_t}{2}\big\Vert \nabla f(\bar{\mathbf{w}}_{t}) \big\Vert^2+\frac{\eta_t}{2N}\sum_{i=1}^N\mathbb{E}\big\Vert  \nabla f(\bar{\mathbf{w}}_{t})\\
    & +\frac{1}{N}\sum_{j=1}^N\big(\nabla f_j(\mathbf{w}_t^j)-\nabla f_j(\mathbf{w}_t^j)  - \nabla F_j(\mathbf{w}_t^j)\big)\big\Vert^2 \\
    &+\frac{\eta_t}{2N}\sum_{i=1}^N\mathbb{E}\big\Vert\sum_{j=1}^N\big(\frac{1}{N}\nabla F_j(\mathbf{w}_t^j)-\theta^{i,j}_t\nabla F_j(\mathbf{w}_t^j) \odot\mathbf{m}_t^{i,j}\big)\big\Vert^2\\
    &\leq  -\frac{\eta_t}{2}\big\Vert \nabla f(\bar{\mathbf{w}}_{t}) \big\Vert^2+\frac{\eta_t\beta^2}{2N}\sum_{i=1}^N\big\Vert \bar{\mathbf{w}}_{t}-\mathbf{w}_t^i\big\Vert^2+\frac{\eta_t(\xi^2+\tau)}{2},
     \end{aligned}
 \end{equation} 
     where $\nabla f(\mathbf{\bar{W}}_t)=[\nabla f(\bar{\mathbf{w}}_{t}),\nabla f(\bar{\mathbf{w}}_{t}),...,\nabla f(\bar{\mathbf{w}}_{t})] \in \mathbb{R}^{d\times N}$. The last inequality holds since Assumption~\ref{smoothness},~\ref{variance} and~\ref{neighborbounded}.

 Then we concentrate on bounding  $T_2$.
 \begin{equation}
     \begin{aligned}
T_2&\leq\frac{\beta\eta_t^2}{2N}\sum_{i=1}^N\mathbb{E}\big\Vert\sum_{j=1}^N\big(\theta^{i,j}_t\nabla F_j(\mathbf{w}_t^j)\odot\mathbf{m}_t^{i,j}- \frac{1}{N}\nabla F_j(\mathbf{w}_t^j)\big)\\
&\quad +\frac{1}{N}\sum_{j=1}^N \big( \nabla F_j(\mathbf{w}_t^j)- \nabla f_j(\mathbf{w}_t^j)+\nabla f_j(\mathbf{w}_t^j)\\
&\quad \quad -\nabla f_j(\bar{\mathbf{w}}_{t})+ \nabla f_j(\bar{\mathbf{w}}_{t})\big)\big\Vert^2\\
&\leq \frac{\beta\eta_t^2}{2}(\tau+\xi^2+\big\Vert \nabla f(\bar{\mathbf{w}}_{t}) \big\Vert^2)+\frac{\beta^3\eta_t^2}{2N}\sum_{i=1}^N\big\Vert \bar{\mathbf{w}}_{t}-\mathbf{w}_t^i\big\Vert^2.
     \end{aligned}
 \end{equation}

%  &\leq\frac{1}{N}\sum_{i=1}^N \mathbb{E}\big\Vert \bar{\mathbf{w}}_{t-1}-\mathbf{w}_{t-1}^i\big\Vert^2+\frac{\eta_{t-1}^2(N+1)}{N}\tau\\

 The term $T_3:=\frac{1}{N}\sum_{i=1}^N\mathbb{E}\big\Vert \bar{\mathbf{w}}_{t}-\mathbf{w}_t^i\big\Vert^2$ can be bounded as:
 \begin{equation}
     \begin{aligned}
  T_3  &    \leq   \frac{1}{N} \sum_{i=1}^N\mathbb{E} \big\Vert \bar{\mathbf{w}}_{t-1}-\mathbf{w}_{t-1}^i\big\Vert^2+\frac{\eta_{t-1}^2}{N}\sum_{i=1}^N\big[(1+\frac{1}{N}) \\
    &\mathbb{E}\big\Vert 
\sum_{j=1}^N  \big( \frac{1}{N}\nabla F_j(\mathbf{w}_{t-1}^j) -\theta_{t-1}^{i,j} \nabla F_j(\mathbf{w}_{t-1}^j)\odot \mathbf{m}^{i,j}_{t-1}\big)\big\Vert^2\big]\\
    &\leq\frac{1}{N}\sum_{i=1}^N \mathbb{E}\big\Vert \bar{\mathbf{w}}_{t-1}-\mathbf{w}_{t-1}^i\big\Vert^2+2\eta_{t-1}^2\tau\\
    &\leq \frac{1}{N}\sum_{i=1}^N \mathbb{E}\big\Vert \bar{\mathbf{w}}_{0}-\mathbf{w}_{0}^i\big\Vert^2+2\tau\sum_{l=1}^{t}\eta_{l-1}^2.
     \end{aligned}
 \end{equation}
 \begin{comment}
      \begin{equation}
     \begin{aligned}
    &    \leq   \frac{1}{N} \sum_{i=1}^N\mathbb{E} \big\Vert \bar{\mathbf{w}}_{t-1}-\mathbf{w}_{t-1}^i\big\Vert^2+\frac{\eta_{t-1}^2}{N}\sum_{i=1}^N \mathbb{E}\big\Vert 
    \sum_{j=1}^N\Tilde{\theta}^{i,j}_t\nabla F_j(\mathbf{w}_t^j)\\
    &-\frac{1}{N}\sum_{j=1}^N\nabla F_j(\mathbf{w}_t^j)+\frac{1}{N}\sum_{j=1}^N\nabla F_j(\mathbf{w}_t^j)-\frac{1}{N}\sum_{i,j=1}^N\Tilde{\theta}^{i,j}_t\nabla F_j(\mathbf{w}_t^j)\big\Vert^2\\
    &\leq\frac{1}{N}\sum_{i=1}^N \mathbb{E}\big\Vert \bar{\mathbf{w}}_{t-1}-\mathbf{w}_{t-1}^i\big\Vert^2+\frac{\eta_{t-1}^2(N+1)}{N}\tau\\
    &\leq \frac{1}{N}\sum_{i=1}^N \mathbb{E}\big\Vert \bar{\mathbf{w}}_{0}-\mathbf{w}_{0}^i\big\Vert^2+\sum_{k=0}^{t-1}\frac{\eta_k^2(N+1)}{N}\tau
     \end{aligned}
 \end{equation}
 
 \end{comment}
 Notice that $\bar{\mathbf{w}}_{0}=\mathbf{w}_{0}^i, \forall i \in [N]$, the term $\frac{1}{N}\sum_{i=1}^N\mathbb{E}\big\Vert \bar{\mathbf{w}}_{t}-\mathbf{w}_t^i\big\Vert^2$ can thus be bounded by $2\tau\sum_{l=1}^{t}\eta_{l-1}^2$.

To incorporate the above results and select the stepsize satisfying $\eta_t=\frac{\eta}{\sqrt{T}} \leq \frac{1}{\beta}$,  we  can  immediately derive the below convergence bound of the considered DFL:
 \begin{equation}
     \begin{aligned}
     \min_{0\leq t\leq T}\big\Vert \nabla f(\bar{\mathbf{w}}_{t}) \big\Vert^2 &\leq     \frac{2(f_0- f^*)}{\eta(\sqrt{T}-\beta\eta)}+\frac{\beta\eta+\sqrt{T}}{\sqrt{T}-\beta\eta}\xi^2\\
         &\quad+\frac{(1+\beta^2\eta^2)(\beta\eta+\sqrt{T})}{(\sqrt{T}-\beta\eta)}\tau.
     \end{aligned}
 \end{equation}

\begin{comment}
    
\end{comment}

\end{proof}
%\mathbb{E}\big\Vert\frac{1}{N}\sum_{j=1}^N\nabla F_j(\mathbf{w}_t^j)-\sum_{j=1}^N\Tilde{\theta}^{i,j}_t\nabla F_j(\mathbf{w}_t^j) \big\Vert^2\\

\subsection{Proof of Theorem~\ref{lemm1}}
\begin{proof}
Based on Assumption~\ref{Lipschitz}, $\Bar{H}$ can be bounded as:
    \begin{equation}
        \begin{aligned}
\Bar{H}&\leq\frac{1}{N}\sum_{i=1}^N\mathbb{E}\Vert \sum_{j=1}^N\big(\theta_{t}^{i,j}\nabla F_j(\mathbf{w}^j_t)\odot \mathbf{p}_{t}^{i,j}- \frac{1}{N}\nabla F_j(\mathbf{w}^j_t)\big)\Vert^2\\
&\quad+\frac{1}{N}\sum_{i=1}^N\mathbb{E}\Vert \sum_{j=1}^N\theta_{t}^{i,j}\nabla F_j(\mathbf{w}^j_t)\odot( \mathbf{p}_{t}^{i,j}- \mathbf{m}_{t}^{i,j})\Vert^2\\
&\leq\frac{1}{N}\sum_{i=1}^N\mathbb{E}\Vert \sum_{j=1}^N\big(\theta_{t}^{i,j}\nabla F_j(\mathbf{w}^j_t)\odot \mathbf{p}_{t}^{i,j}- \frac{1}{N}\nabla F_j(\mathbf{w}^j_t)\big)\Vert^2\\
&\quad+\frac{dL^2}{N}\sum_{i=1}^N\sum_{j=1}^N(\theta_t^{i,j})^2 p_{i,j}^t(1-p_{i,j}^t),
        \end{aligned}
    \end{equation}
where the last inequality holds since the expected square norm  $\mathbb{E}\Vert \mathbf{p}_{t}^{i,j}- \mathbf{m}_{t}^{i,j}\Vert^2=dp_{i,j}^t(1-p_{i,j}^t)$.

\end{proof}

\bibliographystyle{ieeetr}
\bibliography{IEEEtran}

\begin{thebibliography}{10}

\bibitem{DBLP:conf/icc/JiL23}
S.~Ji and M.~Li, ``Enhancing deep learning performance of massive {MIMO} {CSI} feedback,'' in {\em {ICC}}, pp.~4949--4954, {IEEE}, 2023.

\bibitem{DBLP:journals/vtm/XiongZNDWW19}
Z.~Xiong, Y.~Zhang, D.~Niyato, R.~Deng, P.~Wang, and L.~Wang, ``Deep reinforcement learning for mobile 5g and beyond: Fundamentals, applications, and challenges,'' {\em {IEEE} Veh. Technol. Mag.}, vol.~14, no.~2, pp.~44--52, 2019.

\bibitem{xiong2018mobile}
Z.~Xiong, Y.~Zhang, D.~Niyato, P.~Wang, and Z.~Han, ``When mobile blockchain meets edge computing,'' {\em IEEE Communications Magazine}, vol.~56, no.~8, pp.~33--39, 2018.

\bibitem{10319759}
T.~Yin, L.~Li, W.~Lin, T.~Ni, Y.~Liu, H.~Xu, and Z.~Han, ``Joint client scheduling and wireless resource allocation for heterogeneous federated edge learning with non-iid data,'' {\em IEEE Transactions on Vehicular Technology}, pp.~1--13, 2023.

\bibitem{9944162}
H.~Zhu, J.~Kuang, M.~Yang, and H.~Qian, ``Client selection with staleness compensation in asynchronous federated learning,'' {\em IEEE Transactions on Vehicular Technology}, vol.~72, no.~3, pp.~4124--4129, 2023.

\bibitem{DBLP:journals/jstsp/YeLL22}
H.~Ye, L.~Liang, and G.~Y. Li, ``Decentralized federated learning with unreliable communications,'' {\em {IEEE} J. Sel. Top. Signal Process.}, vol.~16, no.~3, pp.~487--500, 2022.

\bibitem{DBLP:journals/tnsm/WuWL23}
Z.~Wu, X.~Wu, and Y.~Long, ``Joint scheduling and robust aggregation for federated localization over unreliable wireless {D2D} networks,'' {\em {IEEE} Trans. Netw. Serv. Manag.}, vol.~20, no.~3, pp.~3359--3379, 2023.

\bibitem{DBLP:conf/sensys/JiX022}
S.~Ji, Y.~Xie, and M.~Li, ``Sifall: Practical online fall detection with {RF} sensing,'' in {\em SenSys}, pp.~563--577, {ACM}, 2022.

\bibitem{DBLP:journals/tccn/XiongZLKNLM21}
Z.~Xiong, Y.~Zhang, W.~Y.~B. Lim, J.~Kang, D.~Niyato, C.~Leung, and C.~Miao, ``Uav-assisted wireless energy and data transfer with deep reinforcement learning,'' {\em {IEEE} Trans. Cogn. Commun. Netw.}, vol.~7, no.~1, pp.~85--99, 2021.

\bibitem{yan2023convergence}
Z.~Yan and D.~Li, ``Convergence time optimization for decentralized federated learning with leo satellites via number control,'' {\em IEEE Transactions on Vehicular Technology}, 2023.

\bibitem{chen2023exploring}
Z.~Chen, W.~Yi, and A.~Nallanathan, ``Exploring representativity in device scheduling for wireless federated learning,'' {\em IEEE Transactions on Wireless Communications}, 2023.

\bibitem{DBLP:conf/aistats/BarsBTLK23}
B.~L. Bars, A.~Bellet, M.~Tommasi, E.~Lavoie, and A.~Kermarrec, ``Refined convergence and topology learning for decentralized {SGD} with heterogeneous data,'' in {\em {AISTATS}}, vol.~206 of {\em Proceedings of Machine Learning Research}, pp.~1672--1702, {PMLR}, 2023.

\bibitem{DBLP:journals/tcc/ZengLYZLLN23}
S.~Zeng, Z.~Li, H.~Yu, Z.~Zhang, L.~Luo, B.~Li, and D.~Niyato, ``Hfedms: Heterogeneous federated learning with memorable data semantics in industrial metaverse,'' {\em {IEEE} Trans. Cloud Comput.}, vol.~11, no.~3, pp.~3055--3069, 2023.

\bibitem{DBLP:journals/tsipn/LiuCZ22}
W.~Liu, L.~Chen, and W.~Zhang, ``Decentralized federated learning: Balancing communication and computing costs,'' {\em {IEEE} Trans. Signal Inf. Process. over Networks}, vol.~8, pp.~131--143, 2022.

\bibitem{yan2023performance}
Z.~Yan and D.~Li, ``Performance analysis for resource constrained decentralized federated learning over wireless networks,'' {\em arXiv preprint arXiv:2308.06496}, 2023.

\bibitem{DBLP:journals/iotj/LiHYKLXN22}
Z.~Li, Y.~He, H.~Yu, J.~Kang, X.~Li, Z.~Xu, and D.~Niyato, ``Data heterogeneity-robust federated learning via group client selection in industrial iot,'' {\em {IEEE} Internet Things J.}, vol.~9, no.~18, pp.~17844--17857, 2022.

\bibitem{9951138}
C.~Peng, Q.~Hu, Z.~Wang, R.~W. Liu, and Z.~Xiong, ``Online-learning-based fast-convergent and energy-efficient device selection in federated edge learning,'' {\em IEEE Internet of Things Journal}, vol.~10, no.~6, pp.~5571--5582, 2023.

\bibitem{chen2020convergence}
M.~Chen, H.~V. Poor, W.~Saad, and S.~Cui, ``Convergence time optimization for federated learning over wireless networks,'' {\em IEEE Transactions on Wireless Communications}, vol.~20, no.~4, pp.~2457--2471, 2020.

\bibitem{yahya2023federated}
M.~Yahya, S.~Maghsudi, and S.~Stanczak, ``Federated learning in uav-enhanced networks: Joint coverage and convergence time optimization,'' {\em IEEE Transactions on Wireless Communications}, 2023.

\bibitem{deng2020edge}
S.~Deng, H.~Zhao, W.~Fang, J.~Yin, S.~Dustdar, and A.~Y. Zomaya, ``Edge intelligence: The confluence of edge computing and artificial intelligence,'' {\em IEEE Internet of Things Journal}, vol.~7, no.~8, pp.~7457--7469, 2020.

\bibitem{ren2020accelerating}
J.~Ren, G.~Yu, and G.~Ding, ``Accelerating dnn training in wireless federated edge learning systems,'' {\em IEEE Journal on Selected Areas in Communications}, vol.~39, no.~1, pp.~219--232, 2020.

\bibitem{DBLP:journals/jsac/HuCL23}
C.~Hu, Z.~Chen, and E.~G. Larsson, ``Scheduling and aggregation design for asynchronous federated learning over wireless networks,'' {\em {IEEE} J. Sel. Areas Commun.}, vol.~41, no.~4, pp.~874--886, 2023.

\bibitem{savazzi2020federated}
S.~Savazzi, M.~Nicoli, and V.~Rampa, ``Federated learning with cooperating devices: A consensus approach for massive iot networks,'' {\em IEEE Internet of Things Journal}, vol.~7, no.~5, pp.~4641--4654, 2020.

\bibitem{DBLP:conf/nips/NguyenTL22}
A.~T. Nguyen, P.~H.~S. Torr, and S.~N. Lim, ``Fedsr: {A} simple and effective domain generalization method for federated learning,'' in {\em NeurIPS}, 2022.

\bibitem{roberts2022principles}
D.~A. Roberts, S.~Yaida, and B.~Hanin, {\em The principles of deep learning theory}.
\newblock Cambridge University Press Cambridge, MA, USA, 2022.

\bibitem{de2022mitigating}
A.~B. de~Luca, G.~Zhang, X.~Chen, and Y.~Yu, ``Mitigating data heterogeneity in federated learning with data augmentation,'' {\em arXiv preprint arXiv:2206.09979}, 2022.

\end{thebibliography}
\end{document}